\pdfoutput=1
\documentclass[twocolumn, 8pt]{article}

\usepackage{lipsum}
\usepackage{JAMIA}
\usepackage{amsmath}
\usepackage{amssymb}
\usepackage{subcaption}
\usepackage{tabularx}
\newcolumntype{C}{>{\centering\arraybackslash}X}
\usepackage{algorithmic, algorithm}

\title{A Statistical Approach to Estimate Sample Size of Machine Learning Models}
\author[1]{Dat Phan-Trong}
\author[1]{Sunil Gupta}
\author[1]{Svetha Venkatesh}
\affil[1]{Deakin Applied Artificial Intelligence Initiative, Deakin University, Australia}
\date{}

\begin{document}

\par\noindent\rule[-7pt]{15.5cm}{0.2em}
\begin{strip}
    \begin{minipage}{.88\textwidth}
        \maketitle
        \small
        \abstractSection
        {Sample size determination for machine learning (ML) prediction models is challenging because conventional power analysis typically requires the predictor-outcome relationship and effect structure to be specified a priori. Nonlinear ML models learn complex prediction surfaces that do not admit straightforward analytical power calculations. We propose a framework that approximates nonlinear ML models with localized linear representations and estimates sample size requirements by evaluating statistical power across these local regions.} 
        {A trained model $f_{\text{base}}$ is approximated by a Rectified Linear Unit (ReLU) neural network, yielding a continuous piecewise linear representation that partitions the feature space into locally linear regions. Within each region, local effect sizes ($R^2_l, f^2_l$) are estimated and region-specific statistical power is calculated using the noncentral $F$-distribution for continuous outcomes and large-sample normal approximations for logistic regression. To aggregate these local quantities, we introduce a volume-weighted coverage power metric that accounts for both regional statistical power and the volume of the corresponding feature-space regions. The minimum global sample size is then determined by finding the smallest sample size that achieves a prespecified coverage threshold $\gamma$, subject to the target power $\beta$ and effect-size filtering threshold $\tau_{R^2}$.} 
        {In synthetic experiments with known structural boundaries, required sample size increased with surface complexity and decreased as weak-effect regions were excluded. Empirical evaluation on three UCI datasets demonstrated stable convergence of volume-weighted coverage to target levels and revealed differences in sample size requirements across model architectures.} 
        {The framework decomposes nonlinear prediction surfaces into locally analyzable regions, enabling classical power calculations in settings where global parametric assumptions are inappropriate. Sample size estimates reflect regional effect heterogeneity and partition structure.} 
        {This approach extends traditional power analysis to nonlinear ML models and provides a structured method for estimating sample size requirements in predictive modeling studies.} 
        {sample size estimation; machine learning; statistical power} 
        \par\noindent\rule[-7pt]{15.5cm}{0.2em}
        \hspace{2cm}
    \end{minipage}
\end{strip}

\section*{Introduction}

Sample size determination is a fundamental component of clinical study design because it directly influences statistical power, model reliability, and resource allocation. Classical sample size calculations are typically derived from parametric statistical models, such as linear regression, where effect sizes and power can be expressed analytically through quantities such as the coefficient of determination ($R^2$) and the global $F$-test \cite{cohen2013statistical}. These methods have been widely adopted because they provide a transparent relationship between study objectives, expected effect sizes, and required cohort size. However, although traditional parametric models can incorporate interactions and selected nonlinear effects, these relationships generally need to be specified a priori through a predefined functional form. This can limit their flexibility in biomedical applications where outcomes may depend on complex, high-order, and spatially varying interactions and nonlinear relationships among predictors.

The increasing adoption of machine learning (ML) methods in healthcare reflects their ability to capture complex relationships that may be difficult to specify a priori using traditional statistical models. Statistical models typically require predefined functional forms for main effects and interaction relationships, whereas modern ML models, including support vector machines, random forests, and neural networks, can learn complex nonlinear patterns without explicitly specifying these relationships. While this flexibility has contributed to improved predictive performance across a range of clinical tasks, it also complicates sample size determination because the learned relationships are not readily characterized within conventional analytical frameworks. Consequently, rigorous and practical methodological guidance for determining sample size in ML-based clinical studies remains limited \cite{balki2019sample, figueroa2012predicting}.

Several studies have attempted to address this problem using data-driven or performance-based criteria. Rajput et al. \cite{rajput2023evaluation} proposed empirical guidelines to evaluate sample adequacy by jointly analyzing feature effect sizes and classification accuracy across incremental subsamples, demonstrating that predictive performance often reaches a plateau beyond which additional data provide diminishing returns. Ghasemzadeh et al. \cite{ghasemzadeh2024toward} addressed sample size determination by developing a simulation-based power analysis framework for machine learning models, demonstrating that robust validation schemes like nested $k$-fold cross-validation are essential to prevent optimistic performance bias from underestimating sample size requirements. In addition, Riley et al. \cite{riley2020calculating} emphasized that sample size calculations for prediction models should focus on controlling overfitting and ensuring precise risk estimation rather than relying exclusively on traditional hypothesis-testing frameworks. These criteria are important for developing reliable prediction models, but they do not directly address whether a given sample size provides adequate statistical power across the heterogeneous regions of a nonlinear prediction function. In particular, global criteria such as anticipated model complexity, predictor-to-sample ratios, or overall prediction precision do not account for the possibility that different regions of the feature space may exhibit substantially different effect sizes and therefore require different amounts of information to achieve adequate power. Our approach complements these existing criteria by evaluating statistical power locally across the learned prediction surface and aggregating it through a volume-weighted coverage measure, thereby providing a complementary perspective on sample size requirements for nonlinear ML models.

Despite these advances, most machine learning studies in clinical research continue to be limited by insufficient sample sizes and inadequate validation procedures \cite{balki2019sample}. A key reason is the absence of a framework that connects the flexibility of modern machine learning models with the well-established principles of statistical power analysis. Classical approaches generally require a known parametric model from which effect sizes and degrees of freedom can be derived, whereas machine learning models typically represent complex nonlinear mappings that do not admit a simple analytical form.

To address this gap, we propose a framework that approximates an arbitrary machine learning prediction surface using a neural network with Rectified Linear Unit (ReLU) activations. Because ReLU networks yield a continuous piecewise linear representation, the learned prediction surface can be decomposed into a collection of local linear regions. Within each region, conventional statistical quantities, including local coefficients of determination and $F$-test-based power, can be estimated using established methods. We then introduce a \textbf{volume-weighted coverage power} measure that aggregates these local power estimates according to the volume of their corresponding regions, providing a global measure of statistical power across the prediction surface. The required sample size is determined as the minimum sample size that achieves a prespecified global coverage-power threshold. By combining local linear power analysis with this global coverage criterion, the proposed framework provides a practical approach for determining sample size requirements for nonlinear machine learning models.

The main contributions of this work are: (1) a ReLU-based local power estimation framework, (2) a volume-weighted aggregation strategy for global sample size determination, and (3) empirical evaluation on synthetic and real-world datasets.
\section*{Background: The Global F-Test and the Linearity Assumption}

Classical sample size calculations for continuous outcomes are often based on multiple linear regression models that describe the relationship between a set of predictors $\mathbf{x} = (x_1,\ldots,x_p)$ and an outcome variable $y$:

\begin{equation}
    y = \beta_0 + \beta_1x_1 + \beta_2x_2 + \cdots + \beta_px_p + \epsilon,
\end{equation}

where $\epsilon \sim \mathcal{N}(0,\sigma^2)$ represents measurement noise. Within this framework, the overall predictive contribution of the covariates is commonly assessed through the global null hypothesis

\begin{equation}
    H_0:\beta_1=\beta_2=\cdots=\beta_p=0.
\end{equation}

The statistical power of the corresponding global $F$-test depends on the strength of the relationship between the predictors and the outcome. This relationship is often summarized through Cohen's effect size $f^2$, which can be expressed in terms of the coefficient of determination $R^2$:

\begin{equation}
    f^2 = \frac{R^2}{1-R^2}.
\end{equation}

Given a desired significance level $\alpha$ and target power $(1-\beta)$, the required sample size can be obtained using the non-central $F$ distribution. This approach forms the basis of many widely used sample size calculations for regression-based studies.

A key assumption underlying this formulation is that the relationship between predictors and outcome can be adequately represented by a single global linear model. In many biomedical applications, however, risk surfaces may exhibit complex nonlinear and locally varying relationships that are difficult to specify a priori using conventional statistical models. Under such conditions, a global linear approximation may not fully capture the structure of the underlying relationship. As a result, the estimated effect size and corresponding sample size requirements may not accurately reflect the complexity of the prediction problem.

The framework proposed in this study addresses this limitation by replacing the single global approximation with a collection of local linear approximations. Specifically, a Rectified Linear Unit (ReLU) network is used to approximate the learned prediction surface and partition the feature space into continuous piecewise linear regions. Within each region, conventional effect size estimation and $F$-test based power calculations can be performed using established statistical methodology. These local estimates are subsequently combined to derive a global sample size recommendation that accounts for heterogeneity in the underlying prediction surface. Rather than replacing classical power analysis, our approach seeks to extend its applicability to settings in which the underlying prediction surface is highly complex and nonlinear such as those modelled by machine learning models.

\section*{Materials and Methods}

\subsection*{Study Overview}

We developed a machine learning framework for estimating sample size requirements in settings where the underlying predictor--outcome relationship has a complex and potentially unknown functional form. The proposed approach is motivated by the observation that many machine learning models can represent complex prediction surfaces that are not readily amenable to conventional power analysis. Although flexible models such as Random Forests, Support Vector Machines, and neural networks can capture nonlinear interactions, their learned representations generally do not provide the explicit parametric structure required by classical sample size calculation methods.

The central idea of the proposed framework is to approximate the prediction surface of an arbitrary machine learning model using a Rectified Linear Unit (ReLU) neural network. Because ReLU networks represent continuous piecewise linear functions, the approximated prediction surface can be partitioned into a collection of local linear regions. Within each region, conventional statistical quantities, including local effect sizes and $F$-test based power calculations, can be estimated using established regression methodology. These local estimates are then aggregated to obtain a global sample size recommendation that reflects the heterogeneous structure of the prediction surface.

The proposed framework is evaluated in both synthetic and real-world settings. For synthetic experiments, the underlying data-generating functions were known, allowing the estimated sample size ($N_{est}$) to be compared against a reference sample size ($N_{true}$) derived from the ground-truth function $f_{\text{true}} : \mathcal{D} \rightarrow \mathbb{R}$, from which pilot data $\mathcal{D}$ are sampled. Additional experiments on benchmark datasets were conducted to assess the framework's behavior across different machine learning architectures and data domains. The overall workflow is summarized in Algorithm~\ref{alg:ssc} and schematically illustrated in Figure~\ref{fig:methodology_power_diagram}.

\subsection*{Data Simulation}

To evaluate the proposed framework under controlled conditions, we generated synthetic datasets designed to emulate nonlinear clinical risk surfaces. Each dataset was constructed within a bounded $d$-dimensional feature domain $\mathcal{D} \subset \mathbb{R}^d$.

The underlying data-generating mechanism was defined by a ground-truth function $f_{\text{true}} : \mathcal{D} \rightarrow \mathbb{R}$ constructed using continuous piecewise linear (CPWL) functions. Functional complexity was controlled by varying the number of knots ($k$) per dimension, thereby introducing structured changes in local gradients to simulate nonlinear threshold effects commonly observed in clinical phenomena.

Synthetic pilot dataset observations $\mathcal{D} = \{(\mathbf{x}_i, y_i)\}_{i=1}^{n}$ were generated by sampling feature vectors $\mathbf{x}_i \sim \mathcal{U}(\mathcal{D})$ from the domain and generating responses $y_i = f_{\text{true}}(\mathbf{x}_i) + \epsilon_i$, where $\epsilon_i \sim \mathcal{N}(0, \sigma^2(\mathbf{x}_i))$ represents heteroscedastic noise. This design tests the proposed framework's capability to distinguish systematic structural signal from stochastic noise.

\subsection*{ML-Assisted Sample Size Framework}

We developed a multi-stage procedure to estimate sample size requirements for nonlinear prediction models. The framework consists of four primary components: (1) base model specification, (2) ReLU-based proxy approximation, (3) local linear power estimation, and (4) global sample size estimation.

\subsubsection*{Base Model Specification and Reference Surface Construction}

In practice, the first step of the framework assumes a user has collected pilot data $\mathcal{D} = \{(\mathbf{x}_i, y_i)\}_{i=1}^{n}$ (or trained a preliminary model) and fitted a prediction model $f_{\text{base}}$. Alternatively, when pilot data are simulated or provided in benchmark evaluations, the dataset $\mathcal{D} = \{(\mathbf{x}_i, y_i)\}_{i=1}^{n}$ is generated from an underlying ground-truth target function $f_{\text{true}} : \mathcal{D} \rightarrow \mathbb{R}$ as $y_i = f_{\text{true}}(\mathbf{x}_i) + \epsilon_i$, where
\begin{equation}
    \mathcal{D} = \{(\mathbf{x}_i, y_i)\}_{i=1}^{n},
\end{equation}
where $\mathbf{x}_i \in \mathcal{D} \subset \mathbb{R}^d$ and $y_i \in \mathbb{R}$.

A nonlinear machine learning model $f_{\text{base}}$, parameterized by estimate $\hat{\theta}$, is trained by minimizing empirical risk:
\begin{equation*}
    \hat{\theta} = \arg\min_{\theta} \frac{1}{n} \sum_{i=1}^{n}
    \mathcal{L}\big(f_{\text{base}}(\mathbf{x}_i; \theta), y_i\big),
\end{equation*}
where $\mathcal{L}$ denotes a task-specific loss function (e.g., mean squared error for regression).

The fitted model defines a reference prediction surface
\begin{equation}
    f_{\text{base}}(\mathbf{x}) = \hat{y}.
\end{equation}

Because this learned surface does not possess an explicit global parametric representation amenable to the forms allowed by traditional power calculators, such tools (traditional analytical power calculators) are not directly applicable. We approximate $f_{\text{base}}$ via a continuous piecewise-linear function.

\begin{figure*}[!t]
    \centering
    \includegraphics[width=0.82\textwidth]{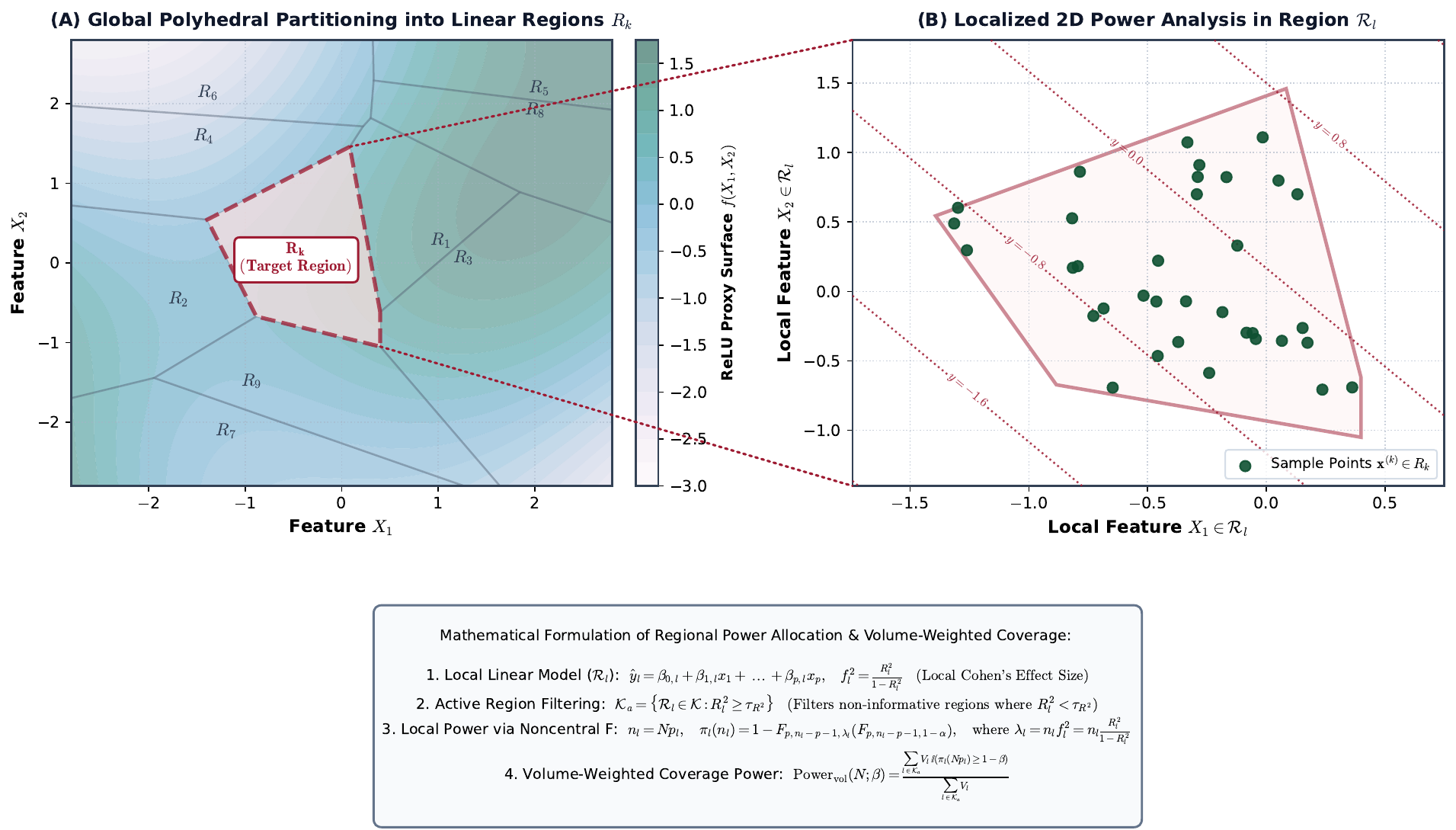}
    \vspace{-0.5ex}
    \caption{Overview of the proposed localized sample size calculation framework in a two-dimensional feature space $(X_1, X_2)$. \textbf{(A) Global Polyhedral Partitioning into Linear Regions:} An arbitrary nonlinear prediction surface is approximated by a continuous piecewise linear (CPWL) ReLU neural network proxy $f(X_1, X_2)$, partitioning the feature domain into convex polyhedral regions $\mathcal{K} = \{\mathcal{R}_1, \dots, \mathcal{R}_L\}$ with a highlighted target region $\mathcal{R}_k$. \textbf{(B) Localized 2D Power Analysis in Region $\mathcal{R}_l$:} Within each polyhedral compartment, the proxy behaves as a local linear regression plane $\hat{y}_l = \beta_{0,l} + \beta_{1,l} x_1 + \dots + \beta_{p,l} x_p$ (indicated by parallel contour lines), populated by local pilot samples $x^{(i)} \in \mathcal{R}_k$. \textbf{Bottom Box:} The four-stage mathematical formulation: (1) estimating local Cohen's effect size $f_l^2 = R_l^2 / (1 - R_l^2)$, (2) filtering out uninformative regions where $R_l^2 < \tau_{R^2}$ to form the active set $\mathcal{K}_a$, (3) evaluating regional statistical power $\pi_l(n_l)$ under expected sample allocation $n_l = N p_l$ via the noncentral $F$-distribution, and (4) computing global volume-weighted coverage power $\text{Power}_{\text{vol}}(N; \beta)$ weighted by geometric volume $V_l$.}
    \label{fig:methodology_power_diagram}
\end{figure*}

\subsubsection*{ReLU Proxy Approximation}

A secondary feedforward neural network with Rectified Linear Unit (ReLU) activations, denoted $f_{\text{ReLU}}$, is trained on the synthetic proxy dataset to approximate the reference prediction surface:
\begin{equation*}
    f_{\text{ReLU}}(\mathbf{x}) \approx f_{\text{base}}(\mathbf{x}).
\end{equation*}

Because ReLU activations are continuous piecewise linear functions, $f_{\text{ReLU}}$ partitions the feature domain into a finite set of convex polyhedral linear regions. To ensure approximation fidelity and avoid geometric shattering, candidate proxy architectures are trained across multiple random initializations, and the network minimizing approximation error (measured by mean squared error) is selected.

\subsubsection*{Local Linearization and Regional Power Allocation}

As illustrated in Figure~\ref{fig:methodology_power_diagram}A, the selected ReLU network induces a polyhedral partition of the feature domain:
\begin{equation}
    \mathcal{K} = \{\mathcal{R}_1, \ldots, \mathcal{R}_L\}.
\end{equation}
Within each polyhedral region $\mathcal{R}_l$, the proxy prediction surface is strictly linear (Figure~\ref{fig:methodology_power_diagram}B):
\begin{equation}
    \hat{y}_l = \beta_{0,l} + \beta_{1,l}x_1 + \cdots + \beta_{p,l}x_p.
\end{equation}

We inherit the polyhedral identification algorithm proposed by Gaines et al. \cite{gaines2026characterizing} to efficiently extract the hyperplane boundaries and regional coefficient vectors $\boldsymbol{\beta}_l = (\beta_{1,l}, \dots, \beta_{p,l})^\top$.

For each region $\mathcal{R}_l$, we compute the local coefficient of determination $R_l^2$ and local Cohen's effect size $f_l^2$:
\begin{equation}
    f_l^2 = \frac{R_l^2}{1 - R_l^2}.
\end{equation}

To filter out negligible local signals, an active region set $\mathcal{K}_a \subseteq \mathcal{K}$ is retained step-by-step based on an effect-size threshold $\tau_{R^2}$:
\begin{equation}
    \mathcal{K}_a = \{\mathcal{R}_l \in \mathcal{K} : R_l^2 \ge \tau_{R^2}\}.
\end{equation}
Regions with $R_l^2 < \tau_{R^2}$ contribute minimal explanatory power and are excluded from the global coverage requirement to avoid mathematically inflating sample size demands for non-informative feature regions.

For a total candidate global sample size $N$, the expected local sample allocation assigned to region $\mathcal{R}_l$ is:
\begin{equation}
    n_l = N p_l,
\end{equation}
where $p_l = \mathbb{P}(\mathbf{x} \in \mathcal{R}_l)$ denotes the empirical probability mass of region $\mathcal{R}_l$. Within region $\mathcal{R}_l$, local statistical power $\pi_l(n_l)$ for detecting the local linear regression signal at significance level $\alpha$ is derived using the noncentral $F$-distribution:
\begin{equation}
    \pi_l(n_l) = 1 - F_{p, n_l - p - 1, \lambda_l}\left(F_{p, n_l - p - 1, 1-\alpha}\right),
\end{equation}
where $\lambda_l = n_l f_l^2 = n_l \frac{R_l^2}{1 - R_l^2}$ is the local noncentrality parameter.

\subsubsection*{Volume-Weighted Coverage Power}

To aggregate regional statistical power into a single global metric, we define volume-weighted coverage power:
\begin{equation}
    \text{Power}_{\text{vol}}(N; \beta) =
    \frac{\sum_{l \in \mathcal{K}_a} V_l \, \mathbb{I}\big(\pi_l(N p_l) \ge 1-\beta\big)}
    {\sum_{l \in \mathcal{K}_a} V_l},
\end{equation}
where:
\begin{itemize}
    \item $\mathcal{K}_a$ is the active region set retained after filtering ($R_l^2 \ge \tau_{R^2}$),
    \item $V_l$ is the geometric volume of region $\mathcal{R}_l$ computed via convex hull integration,
    \item $p_l$ is the empirical probability mass of region $\mathcal{R}_l$,
    \item $\pi_l(n_l)$ is the local statistical power in region $\mathcal{R}_l$,
    \item $1-\beta$ is the targeted local power requirement (e.g., $0.80$),
    \item $\mathbb{I}(\cdot)$ is the indicator function.
\end{itemize}

This metric measures the proportion of the active feature space volume for which the allocated sample size $n_l = N p_l$ achieves local power $\pi_l(n_l) \ge 1-\beta$. \textit{In the special case where the prediction surface is globally linear ($L=1$), this formulation collapses exactly to classical global $F$-test power analysis.}

Local power estimates are aggregated using geometric volume $V_l$ as a weighting factor to emphasize coverage of the prediction surface. Under volume weighting, a region contributes according to the physical feature space volume it occupies, regardless of its sample frequency in pilot data. This formulation is desirable in clinical modeling where low-density feature regions may represent critical high-risk subpopulations or key decision boundaries. Alternative weighting schemes (e.g., probability mass weighting or utility-weighted coverage) can also be substituted depending on clinical priorities.

\subsubsection*{Global Sample Size Optimization}

The optimal global sample size $\hat{N}$ is defined as the minimum candidate sample size required to meet or exceed a target global coverage threshold $\gamma \in (0, 1)$ (e.g., $\gamma = 0.80$ or $0.90$):
\begin{equation}
    \hat{N} =
    \min \left\{ N \in \mathbb{N} :
    \text{Power}_{\text{vol}}(N; \beta) \ge \gamma \right\}.
    \label{eqn:ssc}
\end{equation}

For each region $\mathcal{R}_l$, local power $\pi_l(n_l)$ is a monotonically increasing function of local sample size $n_l$. Because $n_l = N p_l$ scales linearly with $N$, each indicator $\mathbb{I}(\pi_l(N p_l) \ge 1-\beta)$ is non-decreasing in $N$. Consequently, global volume-weighted coverage $\text{Power}_{\text{vol}}(N; \beta)$ is a non-decreasing monotonic function of $N$, guaranteeing that the optimization problem in Eqn.~\ref{eqn:ssc} has a unique, well-defined minimum.

\begin{algorithm*}[ht]
    \caption{ML-Assisted Sample Size Calculation Framework}
    \label{alg:ssc}
    \begin{algorithmic}[1]

        \REQUIRE Pilot dataset $\mathcal{D} = \{(\mathbf{x}_i, y_i)\}_{i=1}^{n}$,
        target local statistical power $1-\beta$,
        target global coverage threshold $\gamma$,
        effect size threshold $\tau_{R^2}$,
        base model $f_{\text{base}}$.
        \ENSURE Globally optimized required sample size $\hat{N}$.

        \STATE \textbf{1. Base Model Surface Specification}
        \STATE Fit base model $f_{\text{base}}$ on pilot data $\mathcal{D}$.

        \STATE \textbf{2. Proxy Approximation and Decomposition}
        \STATE Sample dense domain grid to generate proxy dataset $\mathcal{D}_{\text{proxy}} = \{(\mathbf{x}_i, f_{\text{base}}(\mathbf{x}_i))\}_{i=1}^{n_{\text{proxy}}}$.
        \STATE Train ReLU network $f_{\text{ReLU}}$ to approximate $f_{\text{base}}$.
        \STATE Decompose $f_{\text{ReLU}}$ into polyhedral linear regions $\mathcal{K} = \{\mathcal{R}_1, \dots, \mathcal{R}_L\}$.

        \STATE \textbf{3. Regional Profiling and Active Filtering}
        \STATE Initialize active region set: $\mathcal{K}_a \leftarrow \emptyset$.
        \FOR{each region $\mathcal{R}_l \in \mathcal{K}$}
        \STATE Compute regional coefficient of determination $R_l^2$ on $\mathcal{D}_{\text{proxy}}$.
        \IF{$R_l^2 \ge \tau_{R^2}$}
        \STATE Compute geometric volume $V_l$ of polyhedral region $\mathcal{R}_l$.
        \STATE Estimate empirical probability mass $p_l = \mathbb{P}(\mathbf{x} \in \mathcal{R}_l)$ from pilot data $\mathcal{D}$.
        \STATE Update active set: $\mathcal{K}_a \leftarrow \mathcal{K}_a \cup \{\mathcal{R}_l\}$.
        \ENDIF
        \ENDFOR

        \STATE \textbf{4. Global Sample Size Optimization}
        \STATE Define volume-weighted coverage power:
        \[
            \text{Power}_{\text{vol}}(N; \beta) =
            \frac{\sum_{l \in \mathcal{K}_a}
                V_l \, \mathbb{I}\big(\pi_l(N p_l) \ge 1-\beta\big)}
            {\sum_{l \in \mathcal{K}_a} V_l}.
        \]

        \STATE Solve via monotonic binary search:
        \[
            \hat{N} =
            \min \left\{ N \in \mathbb{N} :
            \text{Power}_{\text{vol}}(N; \beta) \ge \gamma \right\}.
        \]

        \RETURN $\hat{N}$

    \end{algorithmic}
\end{algorithm*}

\section*{Experimental Evaluation}

\subsection*{Synthetic Data Generation}

To evaluate the proposed framework under controlled conditions with a known data-generating mechanism, we constructed multi-dimensional continuous piecewise linear (CPWL) functions with explicitly defined structural boundaries. This design enabled direct comparison between estimated and true local effect structures.

The generative domain was defined as a bounded compact set $\mathcal{D} \subset \mathbb{R}^d$. The domain was partitioned into a structured hyperrectangular grid. Along each coordinate axis $j \in \{1, \dots, d\}$, a sequence of knots $\{\overline{k}_{j,m}\}$ was defined to form an initial uniform mesh. To avoid artificial symmetry, interior knots were perturbed using bounded random displacements:
\begin{equation}
    k_{j,m} = \overline{k}_{j,m} + \delta_{j,m},
\end{equation}
where $\delta_{j,m}$ was sampled from a uniform distribution with bounded support. This perturbation altered local cell geometry while preserving the global domain boundaries.

Scalar values were assigned to each point in the discretized multidimensional feature space and subsequently smoothed using a multidimensional Gaussian filter to reduce extreme local variability. Continuous piecewise linear interpolation over the perturbed grid yielded the target function
\[
    f_{\text{true}} : \mathcal{D} \rightarrow \mathbb{R}.
\]

Synthetic observations were sampled uniformly from the domain,
\[
    \mathbf{x} \sim \mathcal{U}(\mathcal{D}),
\]
and defining responses as
\begin{equation}
    y = f_{\text{true}}(\mathbf{x}) + \epsilon,
\end{equation}
where $\epsilon \sim \mathcal{N}(0, \sigma^2(\mathbf{x}))$ represents heteroscedastic noise. This design tests the proposed framework's capability to distinguish systematic structural signal from stochastic noise.

\subsection*{Real-World Empirical Datasets}

To assess generalizability in empirical settings, we evaluated the framework using three datasets from the UCI Machine Learning Repository representing heterogeneous biomedical and physical modeling tasks.

\begin{itemize}
    \item \textbf{Abalone Dataset ($d = 8$, $n = 4{,}177$):} Predicts the number of shell rings (an age proxy) from physical measurements. The dataset exhibits nonlinear growth patterns and substantial feature collinearity.

    \item \textbf{Concrete Compressive Strength Dataset ($d = 8$, $n = 1{,}030$):} Predicts compressive strength (MPa) from mixture composition and curing age. The relationship between predictors and outcome reflects nonlinear material dynamics.

    \item \textbf{Liver Disorder Dataset ($d = 5$, $n = 345$):} Predicts clinical risk based on liver enzyme concentrations and alcohol consumption. The dataset is characterized by moderate dimensionality and relatively high measurement variability.
\end{itemize}

These datasets differ in dimensionality, signal-to-noise characteristics, and structural smoothness, allowing evaluation across diverse nonlinear prediction regimes.

\subsection*{Baseline ML models and Configuration}

To evaluate robustness across model classes, we considered three commonly used nonlinear regression ML models:

\begin{itemize}
    \item \textbf{Support Vector Regression (SVR):} Implemented with a radial basis function (RBF) kernel to model smooth nonlinear decision surfaces.

    \item \textbf{Random Forest (RF) Regressor:} An ensemble of 100 decision trees, enabling flexible partitioning of the feature space.

    \item \textbf{Multi-Layer Perceptron (MLP):} A fully connected feedforward neural network with ReLU activations, trained using standard backpropagation to approximate nonlinear prediction functions.
\end{itemize}

Hyperparameters were selected using standard validation procedures to ensure stable predictive performance prior to application of the sample size estimation framework.

\section*{Results}

We present the results below, evaluated across (1) controlled synthetic environments with known structural boundaries and (2) empirical benchmark datasets.

\subsection*{Synthetic Evaluation with Known Structural Boundaries}

Synthetic experiments were conducted under explicitly defined continuous piecewise linear (CPWL) data-generating mechanisms. Feature dimensionality was varied ($d \in \{2,3,4,6\}$), and structural complexity was controlled through knot density ($K \in \{2,5,7,13\}$). Pilot sample sizes were set to $n = 500$ for lower-dimensional settings and $n = 5000$ for $d=6$ to ensure stable base model convergence prior to proxy approximation.

\subsubsection*{ReLU Approximation Fidelity}

The ReLU proxy network $f_{\text{ReLU}}$ was trained to approximate the fitted base model surface $f_{\text{base}}$. Approximation quality and stability were assessed by measuring the mean squared error (MSE) relative to $f_{\text{base}}$ across 10 independent random initializations of $f_{\text{ReLU}}$, as well as verifying low variance ($\sigma < 0.05$) in the number of induced polyhedral regions. Across configurations, the proxy consistently achieved stable convergence with low approximation error relative to the base surface, indicating that the induced polyhedral decomposition reflected the geometric structure of the underlying model rather than optimization artifacts.

Because the proxy serves strictly as a geometric representation of the fitted surface (rather than an independent predictive model), this step isolates the topological structure of the learned mapping while preserving its local gradients. The stability observed across random initializations confirms that downstream power estimates are not driven by proxy training instability.

\subsubsection*{Effect Size Filtering and Regional Structure}

Table~\ref{tab:generalized_regional_dynamics} summarizes the number of active regions ($|\mathcal{K}_a|$) and retained hyper-volume under varying $R^2$ thresholds ($\tau_{R^2} \in {0.02, 0.13, 0.26}$), corresponding to small, medium, and large effect sizes based on Cohen's $f^2 \in \{0.02, 0.15, 0.35\}$, respectively. Note that regional counts (e.g., $75.4 \pm 9.0$) are reported as mean values with standard deviations averaged across 10 independent simulation replicates.

\begin{table*}[!t]
    \centering
    \small
    \caption{Aggregated Topological Characteristics Across the Synthetic Benchmarking Matrix. The table reports the mean and standard deviation of the number of active linear regions ($|\mathcal{K}_a|$) alongside the percentage of total domain hyper-volume retained across three sequential clinical effect size thresholds ($\tau_{R^2} \in \{0.02, 0.13, 0.26\}$).}
    \label{tab:generalized_regional_dynamics}
    \begin{tabular}{cclcccccc}
        \hline
                           &                      &                & \multicolumn{3}{c}{\textbf{Active Regions ($|\mathcal{K}_a|$)}} & \multicolumn{3}{c}{\textbf{Mean Volume Retained (\%)}}                                                                                         \\ \cline{4-6} \cline{7-9}
        \textbf{Dim ($d$)} & \textbf{Knots ($K$)} & \textbf{Model} & $\tau_{R^2} = 0.02$                                             & $\tau_{R^2} = 0.13$                                    & $\tau_{R^2} = 0.26$ & $\tau_{R^2} = 0.02$ & $\tau_{R^2} = 0.13$ & $\tau_{R^2} = 0.26$ \\ \hline
        2                  & 2                    & RF             & $9.1 \pm 1.8$                                                   & $5.8 \pm 0.7$                                          & $4.6 \pm 1.2$       & $99.7 \pm 0.3$      & $97.0 \pm 2.6$      & $92.8 \pm 7.2$      \\
        2                  & 2                    & SVM            & $8.7 \pm 1.5$                                                   & $5.9 \pm 1.3$                                          & $4.7 \pm 0.9$       & $99.0 \pm 1.0$      & $95.9 \pm 3.4$      & $91.7 \pm 6.9$      \\
                           &                      & MLP            & $9.7 \pm 1.6$                                                   & $6.9 \pm 1.4$                                          & $5.5 \pm 1.1$       & $99.4 \pm 0.6$      & $96.7 \pm 2.1$      & $93.5 \pm 3.1$      \\
        \hline
                           &                      & RF             & $21.6 \pm 4.6$                                                  & $16.0 \pm 2.6$                                         & $13.0 \pm 2.1$      & $96.2 \pm 5.2$      & $89.4 \pm 7.2$      & $84.7 \pm 8.4$      \\
        2                  & 5                    & SVM            & $19.7 \pm 3.3$                                                  & $15.0 \pm 3.6$                                         & $12.2 \pm 2.6$      & $99.3 \pm 0.8$      & $93.1 \pm 5.9$      & $88.1 \pm 7.1$      \\
                           &                      & MLP            & $20.3 \pm 4.1$                                                  & $14.8 \pm 2.3$                                         & $11.4 \pm 1.9$      & $94.0 \pm 5.1$      & $88.3 \pm 6.9$      & $80.3 \pm 7.3$      \\
        \hline
                           &                      & RF             & $28.2 \pm 5.6$                                                  & $22.3 \pm 3.7$                                         & $18.8 \pm 3.6$      & $98.1 \pm 1.4$      & $92.4 \pm 4.0$      & $82.4 \pm 6.8$      \\
        2                  & 7                    & SVM            & $28.3 \pm 5.3$                                                  & $20.6 \pm 2.9$                                         & $16.6 \pm 2.1$      & $97.7 \pm 1.5$      & $90.6 \pm 3.4$      & $80.5 \pm 5.2$      \\
                           &                      & MLP            & $29.5 \pm 4.7$                                                  & $20.8 \pm 2.2$                                         & $16.0 \pm 1.9$      & $98.9 \pm 0.5$      & $93.3 \pm 2.8$      & $82.7 \pm 5.6$      \\
        \hline
                           &                      & RF             & $99.0 \pm 11.3$                                                 & $74.7 \pm 8.9$                                         & $55.8 \pm 6.2$      & $97.3 \pm 0.9$      & $91.3 \pm 2.8$      & $83.4 \pm 4.2$      \\
        2                  & 13                   & SVM            & $75.4 \pm 9.0$                                                  & $35.5 \pm 4.0$                                         & $23.9 \pm 2.6$      & $94.0 \pm 2.7$      & $71.9 \pm 6.4$      & $56.7 \pm 7.4$      \\
                           &                      & MLP            & $100.3 \pm 11.1$                                                & $69.7 \pm 5.0$                                         & $50.4 \pm 5.1$      & $97.3 \pm 1.0$      & $91.5 \pm 2.0$      & $82.7 \pm 6.1$      \\
        \hline
                           &                      & RF             & $173.1 \pm 20.5$                                                & $116.4 \pm 15.6$                                       & $76.3 \pm 11.1$     & $94.8 \pm 1.4$      & $79.1 \pm 3.4$      & $62.5 \pm 3.9$      \\
        3                  & 6                    & SVM            & $169.9 \pm 19.7$                                                & $118.2 \pm 11.3$                                       & $78.9 \pm 10.9$     & $95.7 \pm 1.1$      & $82.8 \pm 4.2$      & $65.7 \pm 5.4$      \\
                           &                      & MLP            & $165.0 \pm 23.5$                                                & $125.4 \pm 15.6$                                       & $85.1 \pm 9.1$      & $95.5 \pm 1.2$      & $83.1 \pm 4.1$      & $66.0 \pm 6.5$      \\
        \hline
                           &                      & RF             & $89.9 \pm 27.9$                                                 & $70.7 \pm 21.8$                                        & $42.8 \pm 12.3$     & $98.1 \pm 0.8$      & $89.7 \pm 2.9$      & $61.9 \pm 9.0$      \\
        4                  & 4                    & SVM            & $89.8 \pm 17.7$                                                 & $75.1 \pm 14.0$                                        & $55.7 \pm 10.4$     & $98.1 \pm 0.9$      & $92.6 \pm 2.9$      & $80.1 \pm 4.5$      \\
                           &                      & MLP            & $94.7 \pm 14.0$                                                 & $80.0 \pm 9.8$                                         & $56.3 \pm 7.3$      & $98.3 \pm 0.3$      & $94.4 \pm 2.3$      & $77.8 \pm 9.6$      \\
        \hline
    \end{tabular}
\end{table*}

Across all configurations in Table~\ref{tab:generalized_regional_dynamics}, increasing $\tau_{R^2}$ systematically reduces both the number of retained active regions and the total retained volume. This behavior is expected because regions with small $R^2_l$ values contribute limited explanatory power and are therefore excluded from global coverage calculations.

For example, in Table~\ref{tab:generalized_regional_dynamics} ($d=2, K=13$), the SVM model exhibits a reduction from $75.4 \pm 9.0$ active regions at $\tau_{R^2}=0.02$ to $23.9 \pm 2.6$ at $\tau_{R^2}=0.26$, accompanied by a corresponding reduction in retained volume from $94.0\%$ to $56.7\%$. This reflects selective exclusion of low-signal regions rather than instability in the decomposition procedure.

Importantly, the monotonic reduction in both $|\mathcal{K}_a|$ and retained volume across thresholds in Table~\ref{tab:generalized_regional_dynamics} confirms that the filtering mechanism behaves consistently across dimensionalities and model classes.

\subsubsection*{Global Sample Size Optimization}

Global sample size requirements $\hat{N}$ were computed to achieve volume-weighted coverage $\gamma \ge 0.90$ at local power $1-\beta = 0.80$. Detailed optimization results and empirical power convergence metrics are presented in Table~\ref{tab:sample_size_dynamics}. Three consistent patterns emerge from Table~\ref{tab:sample_size_dynamics}:

\begin{table*}[!t]
    \centering
    \small
    \caption{Optimized Global Sample Size Requirements ($\hat{N}$) and Empirical Power Convergence Metrics ($\mu \pm \sigma$) accross synthetic linear piecewise benchmarking configurations.}
    \label{tab:sample_size_dynamics}
    \begin{tabular}{cclccc}
        \hline
                           &                      &                & \multicolumn{3}{c}{\textbf{Optimized Global Sample Size Requirement $\hat{N}$ (Power: $\mu \pm \sigma$)}}                                                                            \\ \cline{4-6}
        \textbf{Dim ($d$)} & \textbf{Knots ($K$)} & \textbf{Model} & $\tau_{R^2} = 0.02$ (Small Effect)                                                                        & $\tau_{R^2} = 0.13$ (Medium Effect) & $\tau_{R^2} = 0.26$ (Large Effect) \\ \hline
        2                  & 2                    & RF             & $252 \,\, (0.90 \pm 0.08)$                                                                                & $207 \,\, (0.90 \pm 0.08)$          & $156 \,\, (0.92 \pm 0.07)$         \\
        2                  & 2                    & SVM            & $343 \,\, (0.90 \pm 0.08)$                                                                                & $186 \,\, (0.90 \pm 0.08)$          & $136 \,\, (0.91 \pm 0.07)$         \\
                           &                      & MLP            & $277 \,\, (0.90 \pm 0.03)$                                                                                & $222 \,\, (0.91 \pm 0.05)$          & $136 \,\, (0.92 \pm 0.07)$         \\
        \hline
                           &                      & RF             & $1116 \,\, (0.90 \pm 0.07)$                                                                               & $585 \,\, (0.91 \pm 0.05)$          & $385 \,\, (0.90 \pm 0.05)$         \\
        2                  & 5                    & SVM            & $1537 \,\, (0.90 \pm 0.05)$                                                                               & $691 \,\, (0.90 \pm 0.04)$          & $330 \,\, (0.90 \pm 0.05)$         \\
                           &                      & MLP            & $1196 \,\, (0.90 \pm 0.04)$                                                                               & $565 \,\, (0.90 \pm 0.05)$          & $365 \,\, (0.91 \pm 0.07)$         \\
        \hline
                           &                      & RF             & $1322 \,\, (0.90 \pm 0.05)$                                                                               & $721 \,\, (0.90 \pm 0.05)$          & $480 \,\, (0.90 \pm 0.08)$         \\
        2                  & 7                    & SVM            & $1953 \,\, (0.90 \pm 0.05)$                                                                               & $906 \,\, (0.90 \pm 0.05)$          & $520 \,\, (0.91 \pm 0.06)$         \\
                           &                      & MLP            & $1732 \,\, (0.90 \pm 0.04)$                                                                               & $1116 \,\, (0.91 \pm 0.03)$         & $470 \,\, (0.90 \pm 0.04)$         \\
        \hline
                           &                      & RF             & $7881 \,\, (0.90 \pm 0.03)$                                                                               & $3815 \,\, (0.90 \pm 0.02)$         & $2075 \,\, (0.90 \pm 0.02)$        \\
        2                  & 13                   & SVM            & $22598 \,\, (0.90 \pm 0.04)$                                                                              & $2855 \,\, (0.90 \pm 0.04)$         & $1335 \,\, (0.90 \pm 0.06)$        \\
                           &                      & MLP            & $8376 \,\, (0.90 \pm 0.02)$                                                                               & $3055 \,\, (0.90 \pm 0.02)$         & $1825 \,\, (0.90 \pm 0.03)$        \\
        \hline
                           &                      & RF             & $32781 \,\, (0.90 \pm 0.02)$                                                                              & $10990 \,\, (0.90 \pm 0.02)$        & $5620 \,\, (0.90 \pm 0.03)$        \\
        3                  & 6                    & SVM            & $29806 \,\, (0.90 \pm 0.02)$                                                                              & $9910 \,\, (0.90 \pm 0.02)$         & $5320 \,\, (0.90 \pm 0.03)$        \\
                           &                      & MLP            & $23166 \,\, (0.90 \pm 0.03)$                                                                              & $11175 \,\, (0.90 \pm 0.03)$        & $5685 \,\, (0.90 \pm 0.03)$        \\
        \hline
                           &                      & RF             & $12249 \,\, (0.90 \pm 0.04)$                                                                              & $6422 \,\, (0.90 \pm 0.05)$         & $3941 \,\, (0.90 \pm 0.05)$        \\
        4                  & 4                    & SVM            & $9513 \,\, (0.90 \pm 0.04)$                                                                               & $5711 \,\, (0.90 \pm 0.03)$         & $3496 \,\, (0.90 \pm 0.03)$        \\
                           &                      & MLP            & $8622 \,\, (0.90 \pm 0.03)$                                                                               & $6272 \,\, (0.90 \pm 0.03)$         & $3786 \,\, (0.90 \pm 0.03)$        \\
        \hline
    \end{tabular}
\end{table*}

\begin{itemize}
    \item \textbf{Dependence on Effect Size Threshold}: Across all configurations in Table~\ref{tab:sample_size_dynamics}, $\hat{N}$ decreases as $\tau_{R^2}$ increases. This relationship follows directly from the local noncentrality parameter,
          \begin{equation}
              \lambda_l = n_l \frac{R^2_l}{1 - R^2_l} = n_l f_l^2.
          \end{equation}

          Regions with small $R^2_l$ require disproportionately large local allocation $n_l = N p_l$ to achieve local power $\pi_l(n_l) \ge 1-\beta$. When such regions are retained (e.g., $\tau_{R^2}=0.02$), the global optimization must allocate sufficient total sample size $N$ so that even weak-signal regions satisfy the local power constraint. When these regions are excluded (e.g., $\tau_{R^2}=0.26$), required sample sizes decrease accordingly.

          For instance, in the $d=2, K=2$ RF configuration in Table~\ref{tab:sample_size_dynamics}, $\hat{N}$ decreases from 252 to 156 when moving from $\tau_{R^2}=0.02$ to $\tau_{R^2}=0.26$. This magnitude of reduction is consistent with the nonlinear dependence of $\lambda_l$ on $R^2_l$.

          Extremely large requirements observed in highly fragmented settings (e.g., $\hat{N}=22{,}598$ for $d=2, K=13$, SVM, $\tau_{R^2}=0.02$ in Table~\ref{tab:sample_size_dynamics}) arise when many low-effect regions are retained. In such cases, the optimization must ensure adequate power simultaneously across a large number of small-probability regions, which mathematically inflates the global allocation. These values reflect the strict coverage constraint ($\gamma \ge 0.90$) rather than numerical instability.

    \item \textbf{Structural Complexity and Sample Size}

          For fixed $\tau_{R^2}$, increasing knot density $K$ increases $\hat{N}$ (Table~\ref{tab:sample_size_dynamics}). For example, under $\tau_{R^2}=0.26$ for RF at $d=2$, $\hat{N}$ increases from 156 ($K=2$) to 2,075 ($K=13$).

          This pattern reflects growth in the number of active regions and corresponding fragmentation of probability mass $p_l$. As $K$ increases, the expected allocation $n_l = N p_l$ becomes smaller for each region unless total $N$ increases proportionally. Thus, higher structural complexity requires larger total sample sizes to maintain local power constraints across active compartments.

    \item \textbf{Architectural Differences}

          Differences across RF, SVM, and MLP models in Table~\ref{tab:sample_size_dynamics} are attributable to how each architecture partitions the feature space. Tree-based models tend to induce localized hyperrectangular partitions, while kernel-based or neural architectures produce smoother polyhedral decompositions.
\end{itemize}
These architectural differences affect: (a) the number of active regions $|\mathcal{K}_a|$, (b) the distribution of regional probability mass $p_l$, and (c) the retained volume under filtering. Importantly, observed differences in $\hat{N}$ across architectures reflect structural characteristics of the learned surfaces rather than differences in optimization tolerance. Across all configurations in Table~\ref{tab:sample_size_dynamics}, empirical coverage at $\hat{N}$ consistently converges to target coverage ($\approx 0.90$) with small variance.

\subsubsection*{Power Curve Dynamics}

\begin{figure*}[htbp]
    \centering

    \noindent
    \makebox[0.328\textwidth][c]{\textbf{\normalsize RF}}\hfill
    \makebox[0.328\textwidth][c]{\textbf{\normalsize SVM}}\hfill
    \makebox[0.328\textwidth][c]{\textbf{\normalsize MLP}}
    \par\vspace{0.8ex}

    \par\noindent\makebox[\textwidth][l]{\textbf{\normalsize \textsf{A.} Small Effect Threshold ($\tau_{R^2} = 0.02$)}}\\[0.4ex]
    \begin{subfigure}[b]{0.328\textwidth}
        \centering
        \includegraphics[width=\textwidth]{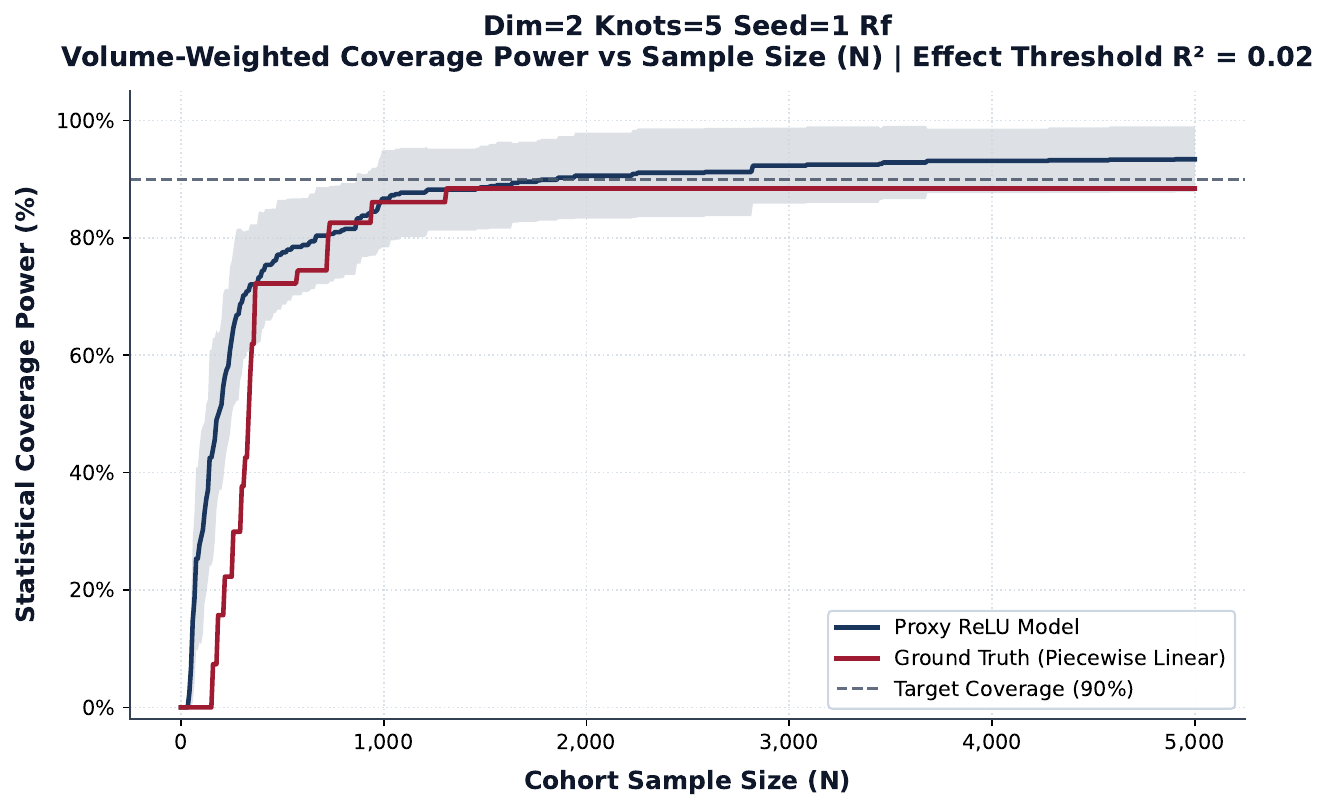}
        \label{fig:rf_02}
    \end{subfigure}
    \hfill
    \begin{subfigure}[b]{0.328\textwidth}
        \centering
        \includegraphics[width=\textwidth]{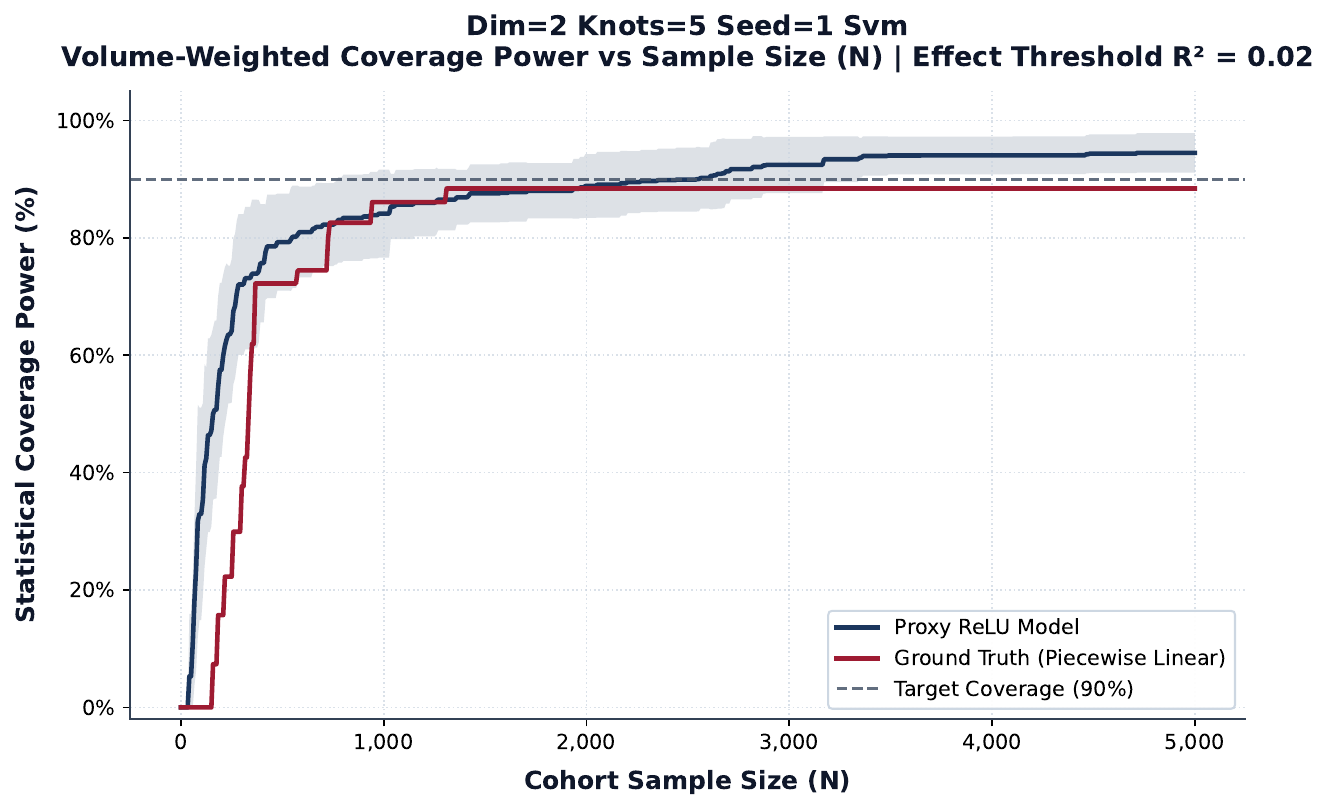}
        \label{fig:svm_02}
    \end{subfigure}
    \hfill
    \begin{subfigure}[b]{0.328\textwidth}
        \centering
        \includegraphics[width=\textwidth]{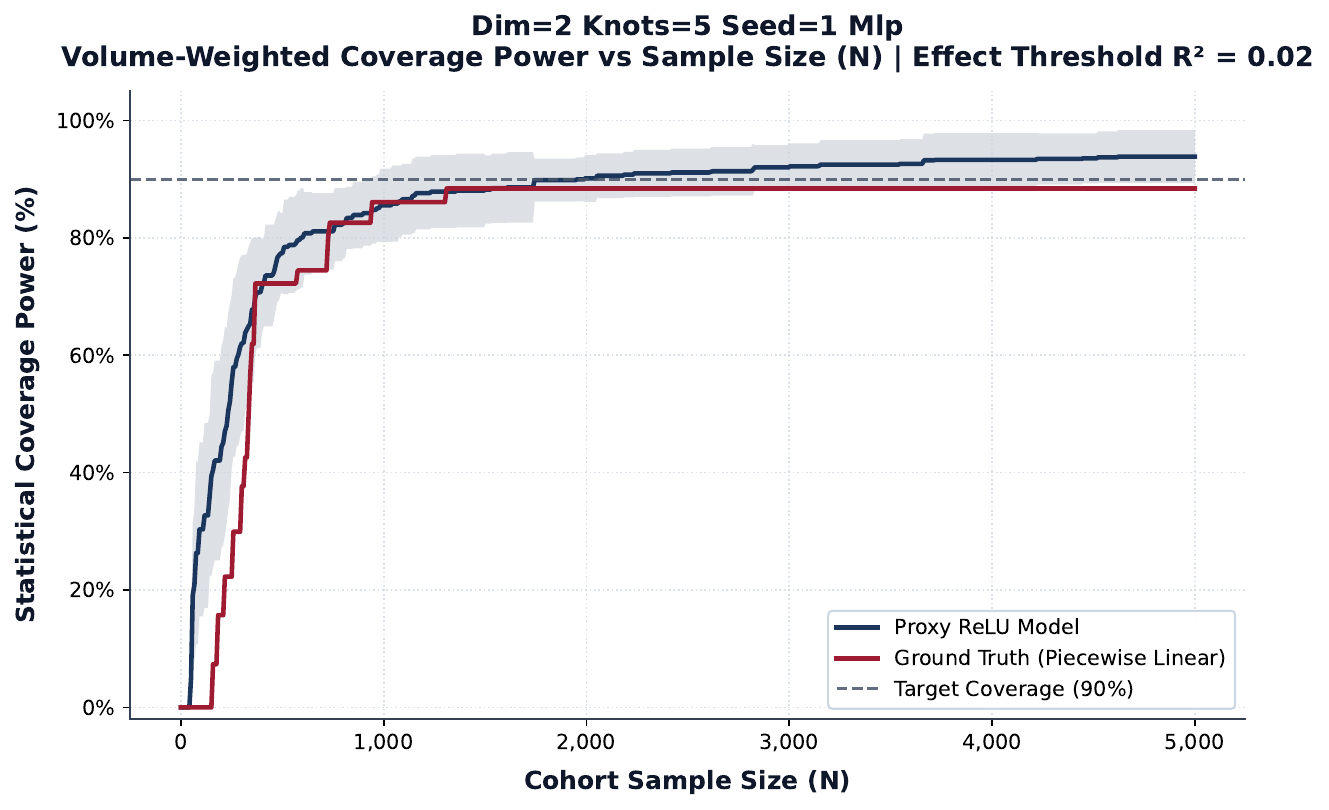}
        \label{fig:mlp_02}
    \end{subfigure}
    \\[1.2ex]

    \par\noindent\makebox[\textwidth][l]{\textbf{\normalsize \textsf{B.} Medium Effect Threshold ($\tau_{R^2} = 0.13$)}}\\[0.4ex]
    \begin{subfigure}[b]{0.328\textwidth}
        \centering
        \includegraphics[width=\textwidth]{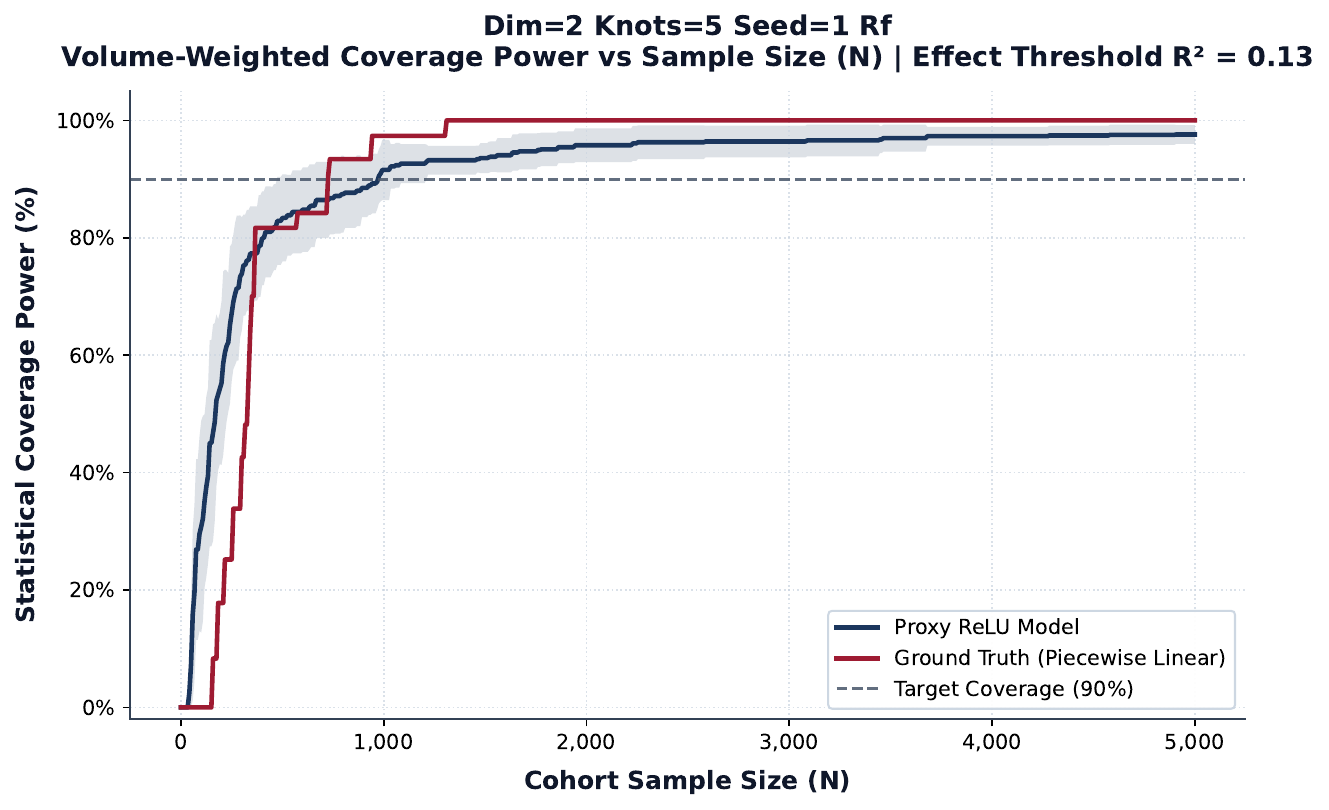}
        \label{fig:rf_13}
    \end{subfigure}
    \hfill
    \begin{subfigure}[b]{0.328\textwidth}
        \centering
        \includegraphics[width=\textwidth]{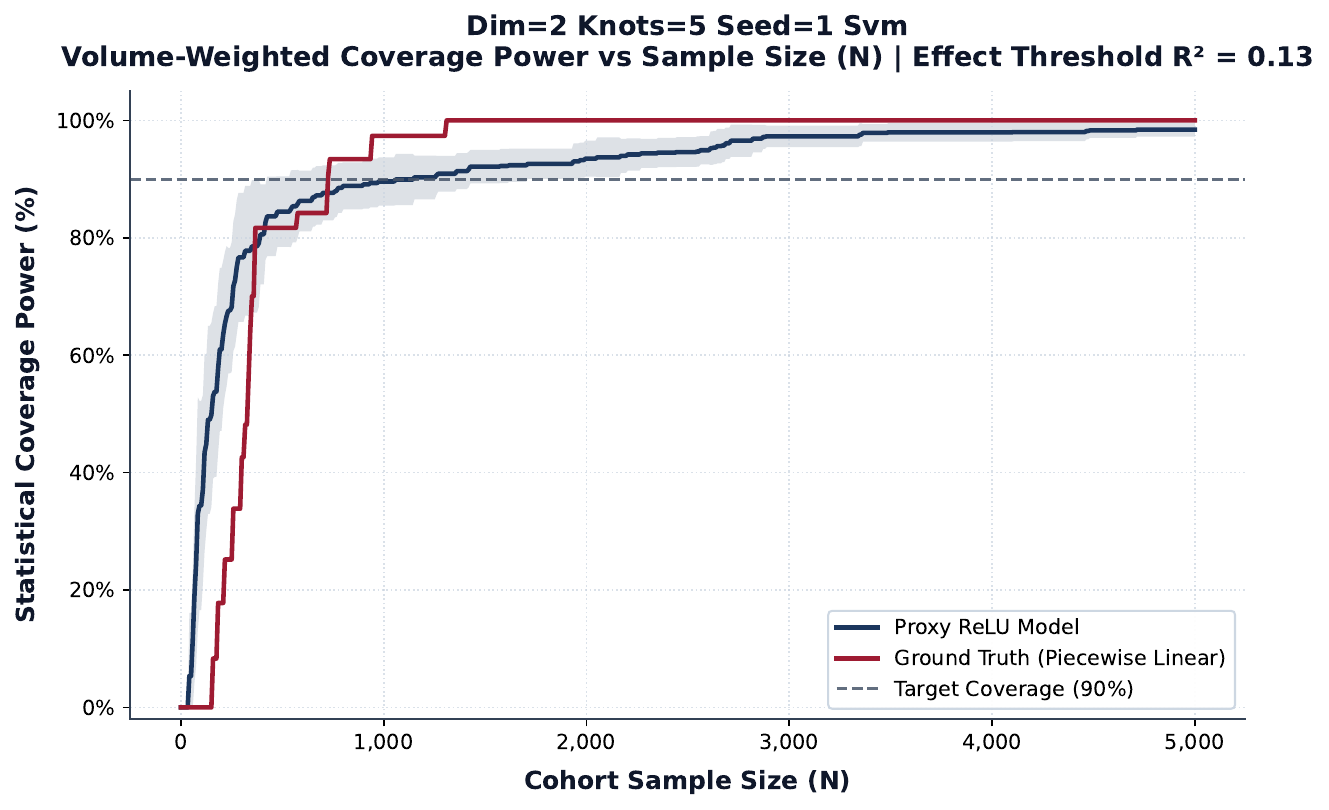}
        \label{fig:svm_13}
    \end{subfigure}
    \hfill
    \begin{subfigure}[b]{0.328\textwidth}
        \centering
        \includegraphics[width=\textwidth]{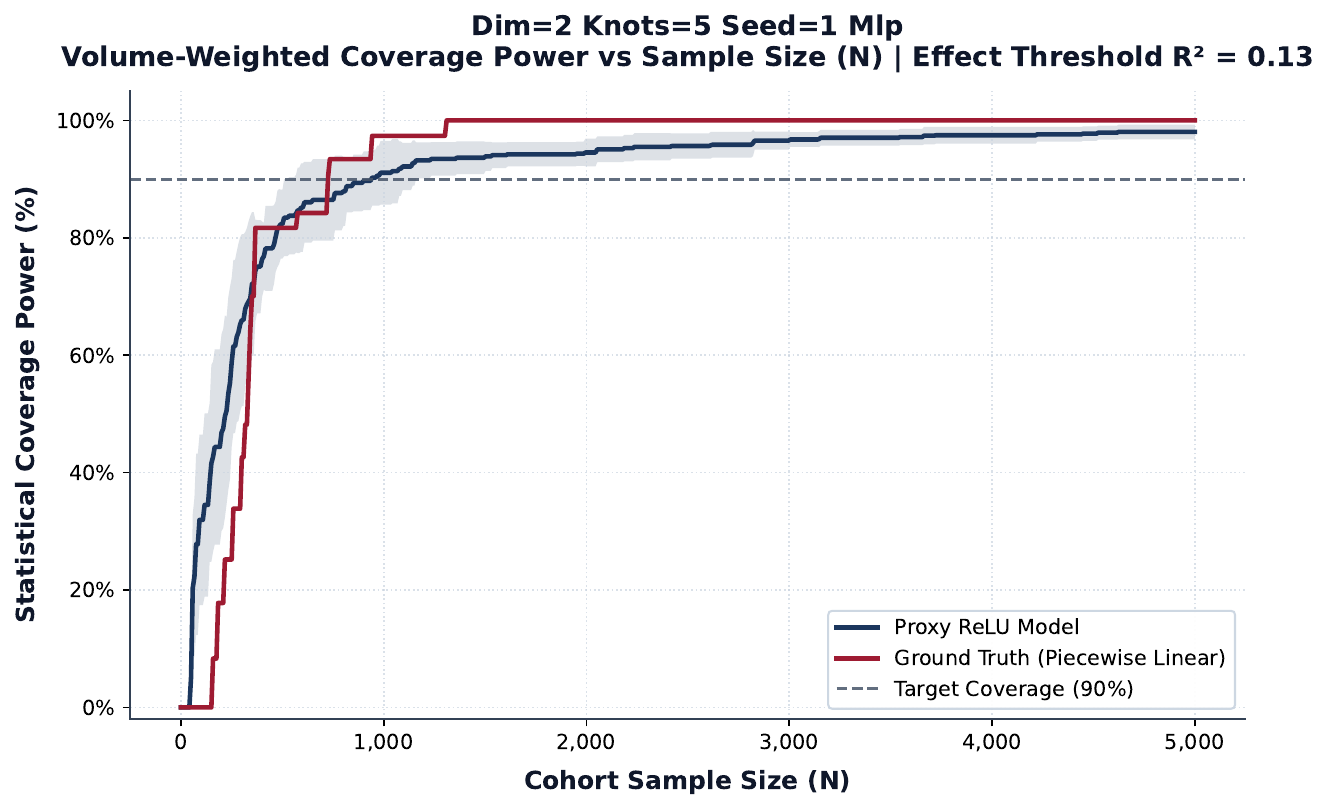}
        \label{fig:mlp_13}
    \end{subfigure}
    \\[1.2ex]

    \par\noindent\makebox[\textwidth][l]{\textbf{\normalsize \textsf{C.} Large Effect Threshold ($\tau_{R^2} = 0.26$)}}\\[0.4ex]
    \begin{subfigure}[b]{0.328\textwidth}
        \centering
        \includegraphics[width=\textwidth]{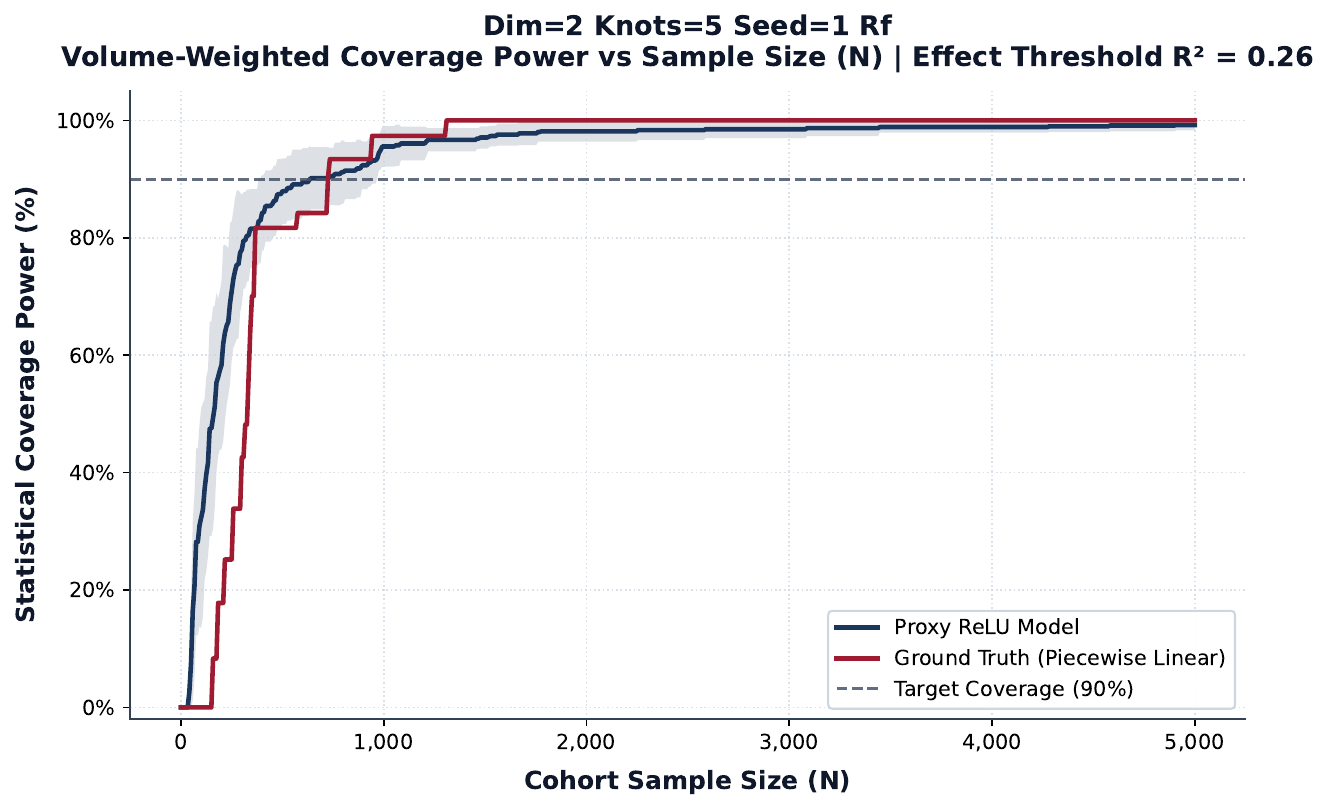}
        \label{fig:rf_26}
    \end{subfigure}
    \hfill
    \begin{subfigure}[b]{0.328\textwidth}
        \centering
        \includegraphics[width=\textwidth]{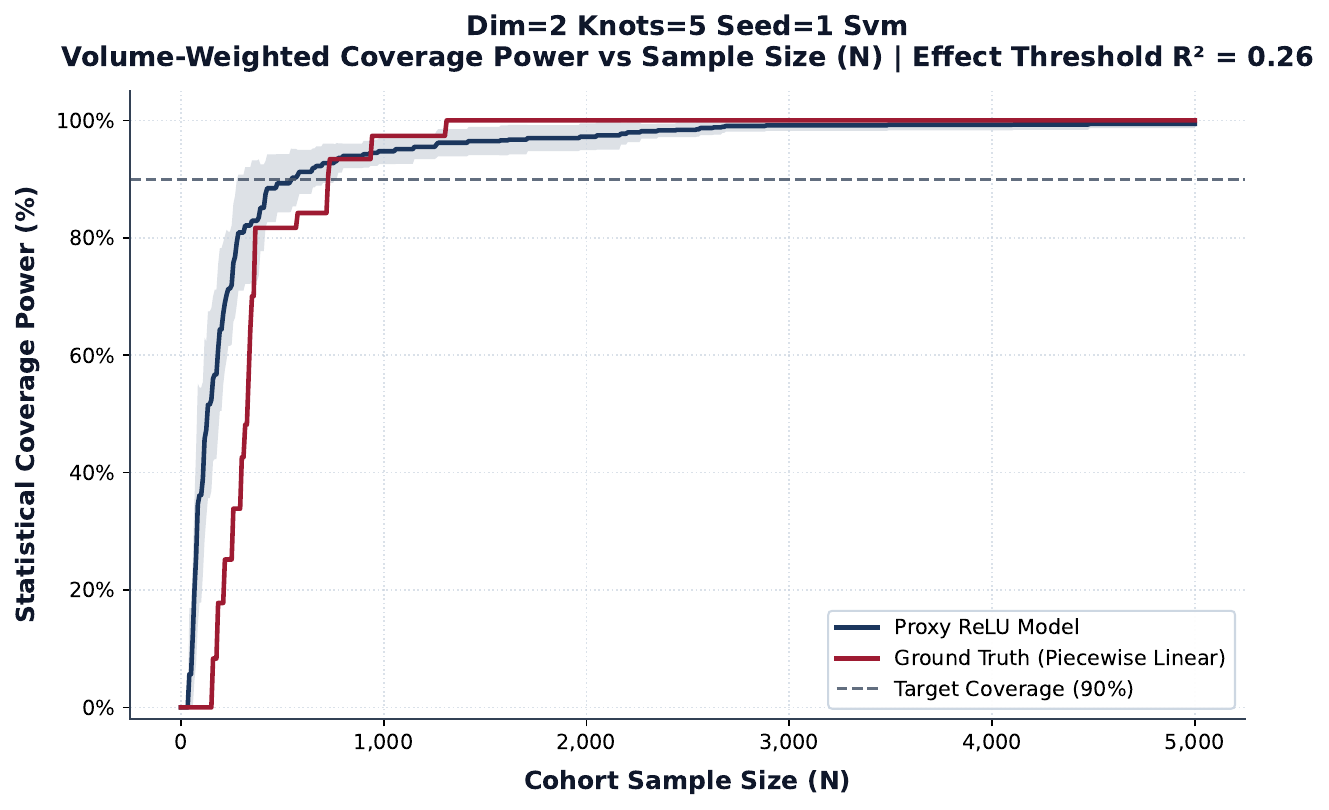}
        \label{fig:svm_26}
    \end{subfigure}
    \hfill
    \begin{subfigure}[b]{0.328\textwidth}
        \centering
        \includegraphics[width=\textwidth]{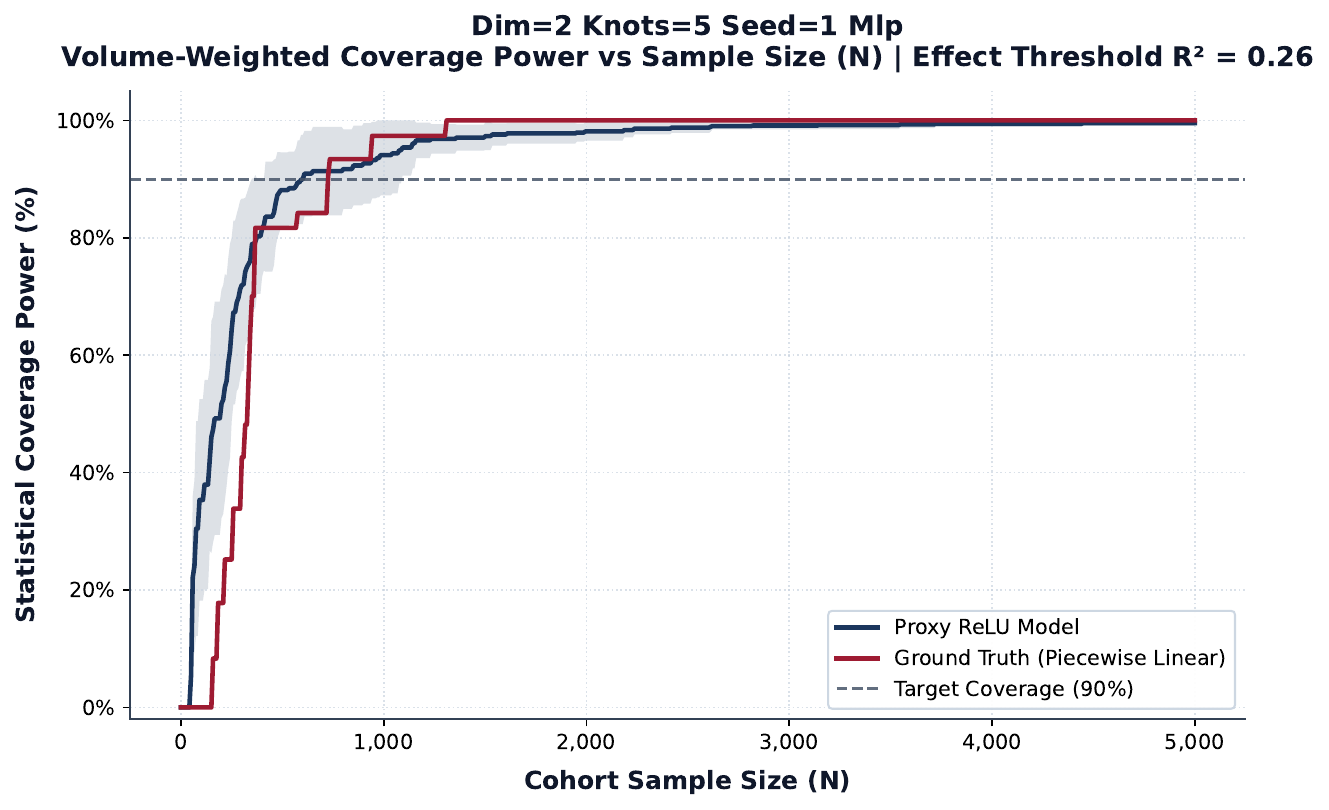}
        \label{fig:mlp_26}
    \end{subfigure}
    \vspace{0.8ex}

    \caption{Volume-weighted coverage power trajectories as a function of total cohort sample size ($N$) for synthetic benchmarking experiments ($d=2, K=5$). \textbf{Columns} represent the three baseline machine learning architectures: Random Forest (RF, left), Support Vector Machine (SVM, center), and Multi-Layer Perceptron (MLP, right). \textbf{Rows} correspond to increasing regional effect-size filtering thresholds: \textbf{(A)} $\tau_{R^2} = 0.02$ (small effect size), \textbf{(B)} $\tau_{R^2} = 0.13$ (medium effect size), and \textbf{(C)} $\tau_{R^2} = 0.26$ (large effect size). In each panel, solid dark-blue curves depict the estimated coverage obtained via the localized ReLU proxy framework (with shaded 95\% empirical variance bands across 10 random proxy initializations), solid dark-red curves denote the ground-truth piecewise linear (CPWL) reference, and dashed horizontal lines mark the target 90\% global coverage threshold ($\gamma = 0.90$). As $\tau_{R^2}$ increases, low-signal regions are excluded from the active partition $\mathcal{K}_a$, shifting coverage curves leftward and reducing the required global sample size $\hat{N}$ while maintaining rigorous coverage fidelity.}
    \label{fig:power_curve_dim2_knots5}
\end{figure*}

Figure~\ref{fig:power_curve_dim2_knots5} illustrates representative volume-weighted coverage trajectories as a function of total sample size $N$. In all cases, coverage increases monotonically with $N$, as expected from the monotonicity of the noncentral $F$-distribution with respect to $\lambda_l$.

Increasing $\tau_{R^2}$ shifts the coverage curve leftward, reflecting exclusion of low-effect regions. The smooth trajectories and narrow variance bands across proxy initializations confirm that global coverage estimates are numerically stable with respect to proxy approximation variability.

\subsection*{Real-World Empirical Evaluation on UCI Datasets}

We evaluated the framework on three benchmark datasets from the UCI Machine Learning Repository: Abalone ($d=8$), Concrete Compressive Strength ($d=8$), and BUPA Liver Disorders ($d=5$). Unlike the synthetic experiments, real-world data introduce two methodological considerations.

First, no analytical ground truth function $f_{\text{true}}$ is available. Accordingly, the framework operates as a post hoc structural diagnostic applied to fitted prediction models rather than as a validation against a known generating mechanism.

Second, uniform grid sampling is inappropriate in empirical datasets due to non-uniform feature distributions and sparsity in high-dimensional space. To approximate the empirical feature manifold while avoiding extrapolation into empty regions, we estimated a Gaussian Mixture Model (GMM) over the observed feature distribution and sampled candidate points from this fitted density. These samples were used to approximate regional probability masses and compute volume-weighted coverage.

For empirical experiments, the global coverage threshold was fixed at $\gamma = 0.80$, and local power was set to $\beta = 0.80$. Optimized sample size estimates $\hat{N}$ and corresponding empirical coverage values are reported in Table~\ref{tab:sample_size_uci}.

\subsubsection*{Dataset-Specific Observations}

Across datasets, three distinct patterns were observed.

\begin{itemize}

    \item \textbf{Concrete Compressive Strength}

          For the Concrete Strength dataset, $\hat{N}$ remains invariant across all effect size thresholds for each model (e.g., RF: 11,611; SVM: 12,412; MLP: 6,206). This indicates that, under the fitted models, few regions exhibit sufficiently small $R^2_l$ values to be excluded by increasing $\tau_{R^2}$. In other words, regional explanatory strength appears relatively homogeneous across the empirical feature manifold.

          Differences across architectures are attributable to variations in how each model partitions feature space and distributes regional probability mass. For example, the MLP yields smaller $\hat{N}$ than RF and SVM in this dataset, reflecting differences in induced regional fragmentation and corresponding local allocations $n_l = N p_l$. These differences arise from structural characteristics of the learned surfaces rather than from changes in the global coverage constraint.

    \item \textbf{BUPA Liver Disorders}

          In the Liver Disorder dataset, $\hat{N}$ decreases with increasing $\tau_{R^2}$ for RF and SVM models (e.g., SVM: 42,642 to 20,620). This behavior mirrors the synthetic experiments: exclusion of low-effect regions reduces the need to simultaneously power multiple weak-signal compartments.

          The magnitude of $\hat{N}$ relative to the original dataset size reflects the strict requirement that 80\% of the empirical feature manifold achieve local power $\ge 0.80$. In fragmented surfaces with heterogeneous regional signal strength, satisfying this constraint may require substantial global allocation. These values therefore quantify structural heterogeneity under the fitted model rather than implying infeasibility of the modeling task.

          The MLP exhibits minimal variation across thresholds (30,830 to 30,630), suggesting that few regions fall below even the strictest $\tau_{R^2}$ threshold. This indicates comparatively stable regional effect sizes under that architecture.

    \item \textbf{Abalone}

          In the Abalone dataset, architectural differences are more pronounced. The RF model produces constant $\hat{N}=8,008$ across thresholds, indicating limited sensitivity to effect-size filtering under the induced partition.

          By contrast, the SVM model exhibits a reduction from 49,649 to 38,638 as $\tau_{R^2}$ increases, suggesting the presence of multiple low-effect regions that are excluded under stricter thresholds. The MLP produces consistently large $\hat{N}$ values (approximately 50,000), reflecting substantial fragmentation and/or diffuse regional probability mass under that architecture.

          These differences underscore that required sample size is a joint function of (i) the fitted surface geometry, (ii) regional probability mass distribution, and (iii) the imposed global coverage constraint.
\end{itemize}
\subsubsection*{Stability of Empirical Coverage Estimates}

Across all datasets and architectures, empirical coverage at $\hat{N}$ converges to approximately $0.80$ with relatively small variance (Table~\ref{tab:sample_size_uci}). This consistency indicates that the optimization routine reliably identifies the crossing point where volume-weighted coverage meets the specified threshold.

Because empirical integration relies on GMM-sampled feature densities, minor variability reflects stochastic sampling rather than instability in the optimization procedure. The observed variance bands remain narrow relative to the coverage threshold, supporting numerical stability of the estimation process.

\begin{table*}[ht]
    \centering
    \small
    \caption{Optimized Global Sample Size Requirements ($\hat{N}$) and Empirical Power Convergence Metrics ($\mu \pm \sigma$) across UCI Repository Manifolds}
    \label{tab:sample_size_uci}
    \begin{tabular}{>{\centering\arraybackslash}p{1.8cm}clccc}\hline
                         &                    &                & \multicolumn{3}{c}{\textbf{Optimized Global Sample Size Requirement $\hat{N}$ (Power: $\mu \pm \sigma$)}}                                                                            \\ \cline{4-6}
        \textbf{Dataset} & \textbf{Dim ($d$)} & \textbf{Model} & $\tau_{R^2} = 0.02$ (Small Effect)                                                                        & $\tau_{R^2} = 0.13$ (Medium Effect) & $\tau_{R^2} = 0.26$ (Large Effect) \\ \hline
                         &                    & RF             & $43843 \,\, (0.80 \pm 0.05)$                                                                              & $39639 \,\, (0.80 \pm 0.05)$        & $35235 \,\, (0.80 \pm 0.05)$       \\
        Liver                                                                                                                                                                                                                                         \\ Disorder & 5 & SVM & $42642 \,\, (0.80 \pm 0.03)$ & $29629 \,\, (0.80 \pm 0.03)$ & $20620 \,\, (0.80 \pm 0.03)$ \\
                         &                    & MLP            & $30830 \,\, (0.80 \pm 0.03)$                                                                              & $30830 \,\, (0.80 \pm 0.03)$        & $30630 \,\, (0.80 \pm 0.03)$       \\
        \hline
                         &                    & RF             & $11611 \,\, (0.80 \pm 0.03)$                                                                              & $11611 \,\, (0.80 \pm 0.03)$        & $11611 \,\, (0.80 \pm 0.03)$       \\
        Concrete                                                                                                                                                                                                                                      \\ Strength & 8 & SVM & $12412 \,\, (0.80 \pm 0.05)$ & $12412 \,\, (0.80 \pm 0.05)$ & $12412 \,\, (0.81 \pm 0.05)$ \\
                         &                    & MLP            & $6206 \,\, (0.80 \pm 0.02)$                                                                               & $6206 \,\, (0.80 \pm 0.02)$         & $6206 \,\, (0.80 \pm 0.02)$        \\
        \hline
                         &                    & RF             & $8008 \,\, (0.80 \pm 0.09)$                                                                               & $8008 \,\, (0.80 \pm 0.09)$         & $8008 \,\, (0.81 \pm 0.09)$        \\
        Abalone          & 8                  & SVM            & $49649 \,\, (0.80 \pm 0.11)$                                                                              & $38638 \,\, (0.81 \pm 0.10)$        & $38638 \,\, (0.81 \pm 0.10)$       \\
                         &                    & MLP            & $50050 \,\, (0.80 \pm 0.08)$                                                                              & $49649 \,\, (0.80 \pm 0.08)$        & $49649 \,\, (0.80 \pm 0.08)$       \\
        \hline
    \end{tabular}
\end{table*}

\section*{Limitations and Future Work}

\begin{itemize}

    \item \textbf{Dependence on Feature Density Estimation:}
          In empirical settings without a known data-generating mechanism, volume-weighted integration relies on sampling from an estimated feature density. In this study, a Gaussian Mixture Model (GMM) was used to approximate the empirical feature distribution. Although this approach reduces extrapolation into unsupported regions of the feature space, results may be sensitive to density misspecification. Rare but structurally important subregions may be underrepresented if not adequately captured by the mixture model. Future work will investigate alternative density estimators, including nonparametric and adaptive sampling strategies, to improve robustness of regional probability mass estimation.

    \item \textbf{Scalability in High Dimensions:}
          The proxy-based decomposition relies on the piecewise linear structure of ReLU networks. As dimensionality increases, the number of induced linear regions may grow rapidly with network depth and width. This expansion increases both memory and computational demands during regional enumeration and volume integration. Practical deployment in very high-dimensional settings may therefore require architectural constraints, dimensionality reduction, or region-clustering approximations to maintain tractable optimization.

    \item \textbf{Extension to Classification and Alternative Outcome Models:}
          The primary formulation presented in the main text anchors local power estimation on continuous outcomes using classical multiple regression assumptions and noncentral $F$-test mechanics. However, many biomedical and clinical applications involve binary classification outcomes (e.g., disease diagnosis, 30-day readmission, or treatment response).

          To accommodate binary classification tasks, the localized power framework extends directly to multiple logistic regression models by substituting large-sample normal approximations for regional log-odds coefficients. Within each induced polyhedral region $\mathcal{R}_l$, the proxy network defines a locally linear log-odds surface, enabling local sample size requirements to be calculated via regional variance inflation factors ($\text{VIF}_l$) and regional event probabilities ($P_l$). Monotonic volume-weighted coverage optimization is then applied analogously to derive the global sample size $\hat{N}$. A complete mathematical derivation and empirical validation matrix for logistic regression classification tasks are detailed in the Supplementary Material (Appendix S1, Table~\ref{tab:sample_size_dynamics_classification}). Extending localized power mapping to time-to-event endpoints (Cox survival models) and correlated longitudinal structures represents an active direction for future research.

\end{itemize}

\section*{Conclusion}

We present a framework for estimating sample size requirements for nonlinear machine learning prediction models by leveraging the continuous piecewise linear structure of ReLU networks. By decomposing complex prediction surfaces into locally linear regions, the approach enables region-specific power calculations that are subsequently aggregated through a volume-weighted coverage criterion.

Across synthetic and empirical evaluations, the framework produces internally consistent sample size estimates that reflect structural characteristics of the fitted prediction surface, including effect size heterogeneity and regional fragmentation. The method also provides insight into how model architecture influences regional partitioning and corresponding sample size requirements.

Rather than replacing classical power analysis, this approach extends its applicability to nonlinear predictive settings where global parametric assumptions are not appropriate. With further refinement of density estimation and scalability components, the framework may serve as a complementary tool for study planning in settings involving complex machine learning models.

\section*{Funding}
This research was partially supported by the Wellcome Trust  [grant number: 303030/Z/23/Z], the National Health and Medical Research Council (NHMRC) Centre of Research Excellence in Depression Treatment Precision (grant number: 2024796) and the NHMRC Synergy Grant (grant number: 2026505).

\bibliographystyle{vancouver}

\bibliography{citation}

\pagebreak
\onecolumn
\begin{appendix}
    \section*{Supplementary Material}

    \subsection*{Extension to Multiple Logistic Regression Tasks}

    To extend the proposed framework beyond continuous outcomes, we outline an adaptation for multiple logistic regression models with binary responses $Y \in \{0,1\}$. In many clinical applications, interest centers on estimating the effect of a target covariate $X_1$ while adjusting for $p-1$ additional predictors. Classical sample size calculations for multivariable logistic regression are typically derived using large-sample normal approximations and may require iterative procedures or simplifying assumptions \cite{hsieh1998simple}.

    Leveraging the continuous piecewise linear (CPWL) proxy network, the feature space is partitioned into polyhedral regions $\mathcal{K}$. Within each region $l$, the proxy induces a locally linear representation of the log-odds surface. This localized structure enables region-specific application of large-sample approximations for logistic regression power calculations.

    \subsubsection*{Localized Sample Size Estimation}

    Within a given active region $l \in \mathcal{K}$, consider estimation of the coefficient associated with $X_1$. Following the normal approximation described by Hsieh et al.~\cite{hsieh1998simple}, the baseline sample size required to detect a log-odds coefficient $\beta_{1,l}^*$ with two-sided significance level $\alpha$ and target power $1-\beta$ is

    \begin{equation}
        n_{1,l} =
        \frac{(Z_{1-\alpha/2} + Z_{1-\beta})^2}
        {P_l(1 - P_l)\beta_{1,l}^{*2}},
    \end{equation}
    \noindent where $Z_{1-\alpha/2}$ and $Z_{1-\beta}$ denote standard normal quantiles, $\beta_{1,l}^*$ is the localized log-odds slope within region $l$, and $P_l$ is the regional event probability.

    To account for correlation between $X_1$ and the remaining covariates $(X_2, \dots, X_p)$, the variance of the estimated coefficient is inflated by a regional variance inflation factor (VIF) \cite{hsieh1998simple}:

    \begin{equation}
        \text{VIF}_l =
        \frac{1}{1 - \rho_{1.23\dots p,l}^2},
    \end{equation}
    \noindent where $\rho_{1.23\dots p,l}^2$ is the squared multiple correlation coefficient obtained by regressing $X_1$ on $(X_2, \dots, X_p)$ within region $l$.

    Operationally, $\rho_{1.23\dots p,l}^2$ is estimated via localized ordinary least squares (OLS). Let $\text{SSR}_l$ denote the sum of squared residuals from regressing $X_1$ on the remaining covariates within region $l$, and let $\text{SST}_l$ denote the corresponding total sum of squares. Then,

    \begin{equation}
        \rho_{1.23\dots p,l}^2 =
        1 - \frac{\text{SSR}_l}{\text{SST}_l}.
    \end{equation}
    \noindent The multivariable regional sample size requirement becomes

    \begin{equation}
        n_{p,l}
        =
        n_{1,l} \cdot \text{VIF}_l
        =
        \frac{(Z_{1-\alpha/2} + Z_{1-\beta})^2}
        {(1 - \rho_{1.23\dots p,l}^2) P_l(1 - P_l)\beta_{1,l}^{*2}}.
    \end{equation}

    These expressions rely on standard large-sample approximations for logistic regression and are therefore most appropriate when regional sample allocations are not extremely small.

    \subsubsection*{Local and Global Power Aggregation}

    Let $N$ denote a candidate global sample size. Under the empirical feature distribution, each region $l$ has probability mass $p_l = \mathbb{P}(\mathbf{x} \in \mathcal{R}_l)$. The expected allocation within region $l$ is $N_l = N p_l$.

    Using the normal approximation, the localized statistical power for detecting $\beta_{1,l}^*$ can be expressed as

    \begin{equation}
        \pi_l(N)
        =
        \Phi\left(
        \sqrt{
            \frac{N p_l}{\text{VIF}_l}
            P_l(1 - P_l)\beta_{1,l}^{*2}
        }
        -
        Z_{1-\alpha/2}
        \right),
    \end{equation}
    \noindent where $\Phi(\cdot)$ denotes the standard normal cumulative distribution function.

    A region $l$ is considered adequately powered if $\pi_l(N) \ge 1 - \beta$. Global coverage is then defined analogously to the regression case:

    \begin{equation}
        \text{Power}_{\text{vol}}(N; \beta)=\frac{\sum_{l \in \mathcal{K}}w_l \,\mathbb{I}\!\left(\pi_l(N) \ge 1-\beta\right)
        }{\sum_{l \in \mathcal{K}}w_l},
    \end{equation}
    \noindent where $w_l$ denotes the geometric volume of region $l$ and $\mathbb{I}(\cdot)$ is an indicator function.

    The required global sample size is defined as

    \begin{equation}
        \hat{N}
        =
        \min
        \left\{
        N \in \mathbb{N}
        \ \middle| \
        \text{Power}_{\text{vol}}(N; \beta) \ge \gamma
        \right\}.
    \end{equation}

    This extension preserves the localized decomposition and volume-weighted aggregation structure of the regression framework while adapting the power calculation to large-sample logistic regression theory.

    \begin{table}[ht]
        \centering
        \small
        \caption{Optimized Global Sample Size Requirements ($\hat{N}$) and Empirical Power Convergence Metrics ($\mu \pm \sigma$) for Logistic Regression across Synthetic Linear Piecewise Benchmarkings.}
        \label{tab:sample_size_dynamics_classification}
        \begin{tabular}{cclccc}
            \hline
                               &                      &                & \multicolumn{3}{c}{\textbf{Optimized Global Sample Size Requirement $\hat{N}$ [Power: $\mu \pm \sigma$]}}                                                                            \\ \cline{4-6}
            \textbf{Dim ($d$)} & \textbf{Knots ($K$)} & \textbf{Model} & $\tau_{R^2} = 0.02$ (Small Effect)                                                                        & $\tau_{R^2} = 0.13$ (Medium Effect) & $\tau_{R^2} = 0.26$ (Large Effect) \\ \hline
            \hline
                               &                      & RF             & $1615 \,\, (0.83 \pm 0.04)$                                                                               & $1615 \,\, (0.85 \pm 0.07)$         & $1555 \,\, (0.80 \pm 0.11)$        \\
            2                  & 5                    & SVM            & $1245 \,\, (0.81 \pm 0.08)$                                                                               & $1245 \,\, (0.81 \pm 0.08)$         & $1245 \,\, (0.81 \pm 0.08)$        \\
                               &                      & MLP            & $3246 \,\, (0.80 \pm 0.10)$                                                                               & $1435 \,\, (0.83 \pm 0.08)$         & $1175 \,\, (0.81 \pm 0.17)$        \\
            \hline
                               &                      & RF             & $2056 \,\, (0.80 \pm 0.08)$                                                                               & $1820 \,\, (0.81 \pm 0.05)$         & $1595 \,\, (0.81 \pm 0.09)$        \\
            2                  & 7                    & SVM            & $4532 \,\, (0.80 \pm 0.14)$                                                                               & $4532 \,\, (0.80 \pm 0.14)$         & $1920 \,\, (0.80 \pm 0.08)$        \\
                               &                      & MLP            & $1725 \,\, (0.83 \pm 0.09)$                                                                               & $1545 \,\, (0.81 \pm 0.11)$         & $1125 \,\, (0.81 \pm 0.09)$        \\
            \hline
                               &                      & RF             & $555 \,\, (0.81 \pm 0.17)$                                                                                & $420 \,\, (0.80 \pm 0.05)$          & $420 \,\, (0.80 \pm 0.05)$         \\
            2                  & 13                   & SVM            & $845 \,\, (0.81 \pm 0.11)$                                                                                & $845 \,\, (0.81 \pm 0.11)$          & $695 \,\, (0.81 \pm 0.12)$         \\
                               &                      & MLP            & $690 \,\, (0.80 \pm 0.07)$                                                                                & $690 \,\, (0.81 \pm 0.06)$          & $655 \,\, (0.80 \pm 0.08)$         \\
            \hline
        \end{tabular}
    \end{table}

\end{appendix}
\end{document}